\documentclass[11pt]{article}

\usepackage[preprint]{acl}

\usepackage{times}
\usepackage{latexsym}
\usepackage{tikz}
\usetikzlibrary{positioning}
\usepackage{subcaption}
\usepackage{enumitem}
\usepackage{float}
\usepackage{booktabs}
\usepackage{amsmath}
\usepackage[T1]{fontenc}
\usepackage[utf8]{inputenc}
\usepackage{microtype}
\usepackage{inconsolata}
\usepackage{graphicx}
\usepackage{amsfonts}
\usepackage{silence}
\title{Is my model ``mind blurting''? Interpreting the dynamics of reasoning tokens with Recurrence Quantification Analysis (RQA)}

\author{
  Quoc Tuan Pham$^{1}$,
  Mehdi Jafari$^{1}$,
  Flora Salim$^{1}$ \\
  $^{1}$School of Computer Science and Engineering, University of New South Wales, Australia \\
  \texttt{peter.pham1@student.unsw.edu.au} \\
\texttt{\{mehdi.jafari, flora.salim\}@unsw.edu.au}}

\begin{document}
\maketitle

\begin{abstract}
Test-time compute is central to large reasoning models, yet analysing their reasoning behaviour through generated text is increasingly impractical and unreliable. Response length is often used as a brute proxy for reasoning effort, but this metric fails to capture the dynamics and effectiveness of the Chain of Thoughts (CoT) or the generated tokens. We propose Recurrence Quantification Analysis (RQA) as a non-textual alternative for analysing model's reasoning chains at test time. By treating token generation as a dynamical system, we extract hidden embeddings at each generation step and apply RQA to the resulting trajectories. RQA metrics, including Determinism and Laminarity, quantify patterns of repetition and stalling in the model’s latent representations. Analysing 3,600 generation traces from DeepSeek-R1-Distill, we show that RQA captures signals not reflected by response length, but also substantially improves prediction of task complexity by 8\%. These results help establish RQA as a principled tool for studying the latent token generation dynamics of test-time scaling in reasoning models.
The codes are available here: \url{https://github.com/cruiseresearchgroup/RQA-CoT/}. 
\end{abstract}

\section{Introduction}
\begin{figure*}[t]
\centering
\sffamily 
\begin{tikzpicture}[scale=0.88, every node/.style={font=\sffamily\small}]

    \node[anchor=west, font=\sffamily\bfseries\large, blue!60!black] at (0, 6.5) {The Challenges of Analysing Long-Chain Reasoning};

    \node[draw, thick, fill=gray!5, text width=3.4cm, minimum height=3.5cm, rounded corners] (c1) at (0,4.2) {};
    \node[anchor=north, font=\sffamily\bfseries\scriptsize, yshift=-0.1cm] at (c1.north) {1. Audit Bottleneck};
    \node[text width=3cm, align=center, gray!60] at (0,4.2) {
        \tiny \texttt{Token 1...\\Token 100...\\Token 1000...\\Token 10000...}
    };
    \node[anchor=south, font=\sffamily\tiny\itshape, text width=3cm, align=center] at (0,2.5) {Volume exceeds human verification capacity};

    \node[draw, thick, fill=gray!5, text width=3.4cm, minimum height=3.5cm, rounded corners] (c2) at (4,4.2) {};
    \node[anchor=north, font=\sffamily\bfseries\scriptsize, yshift=-0.1cm] at (c2.north) {2. Length $\neq$ Depth};
    \draw[fill=green!20, draw=green!50!black] (2.8, 3.8) rectangle (3.5, 4.0) node[right, font=\sffamily\tiny, black, xshift=0.3cm] {Effective};
    \draw[fill=orange!20, draw=orange!50!black] (2.8, 3.4) rectangle (3.5, 3.6) node[right, font=\sffamily\tiny, black, xshift=0.3cm] {Blurting};
    \node[anchor=south, font=\sffamily\tiny\itshape, text width=3cm, align=center] at (4,2.5) {Compute duration does not guarantee quality};

    \node[draw, thick, fill=gray!5, text width=3.4cm, minimum height=3.5cm, rounded corners] (c3) at (8,4.2) {};
    \node[anchor=north, font=\sffamily\bfseries\scriptsize, yshift=-0.1cm] at (c3.north) {3. Unreliable Text};
    \node[draw, rounded corners, fill=white, font=\sffamily] (speech) at (8, 5.2) {\tiny "Step 5: Correct"};
    \node[circle, draw, inner sep=1pt, fill=white, font=\sffamily] (internal) at (8, 4.2) {\tiny Hidden State};
    \draw[<->, orange!80!black, ultra thick, dashed] (speech) -- (internal) node[midway, right, font=\sffamily\tiny] {Faithfulness Gap};
    \node[anchor=south, font=\sffamily\tiny\itshape, text width=3cm, align=center] at (8,2.5) {Textual CoT is often unfaithful to latent logic};

    \node[draw, thick, fill=gray!5, text width=3.4cm, minimum height=3.5cm, rounded corners] (c4) at (12,4.2) {};
    \node[anchor=north, font=\sffamily\bfseries\scriptsize, yshift=-0.1cm] at (c4.north) {4. Static Analysis Bias};
    \begin{scope}[shift={(11.3,3.5)}]
        \draw[fill=gray!20] (0,0) -- (1.4,0) -- (1.6,0.2) -- (0.2,0.2) -- cycle;
        \draw[thick, black] (0.8, 0.4) circle (0.3cm); 
        \draw[thick] (0.8, 0.1) -- (0.8, -0.1);
    \end{scope}
    \node[anchor=south, font=\sffamily\tiny\itshape, text width=3cm, align=center] at (12,2.5) {Current tools ignore temporal dependencies};

    \node[fill=white, draw=gray!30, rounded corners, font=\sffamily\bfseries\footnotesize] at (6, 1.7) {Solution: Recurrence Quantification Analysis (RQA)};

    \node[draw, blue!80!black, ultra thick, fill=blue!2, text width=15cm, minimum height=4.3cm, rounded corners] (solution) at (6.8,-1.8) {};
    
    \begin{scope}[shift={(-0.5,-2.5)}]
        \node[anchor=west, font=\sffamily\bfseries] at (-2, 3.5) {Step 1: Extract Latent Trajectory};
        \draw[->, ultra thick, blue!60!black, opacity=0.3] (0.5,0.5) .. controls (1,2) and (2,0) .. (2.5,1) .. controls (3,2) and (2.5, 3) .. (2, 2.5);
        \draw[<->, thick, blue!40!black, dashed] (1.3, 1.1) -- (2.2, 2.4) node[midway, right, font=\sffamily\tiny\bfseries, black] {Recurrence};
        \node[font=\sffamily\tiny, text width=4cm] at (1.5, 0) {Treats hidden states as a continuous dynamical system.};
    \end{scope}

    \begin{scope}[shift={(5.5,-2.5)}]
        \node[anchor=west, font=\sffamily\bfseries] at (0, 3.5) {Step 2: Map Dynamics};
        \draw[thick, fill=white] (0.5,0.5) rectangle (2.5,2.5);
        \foreach \i in {0.6, 1.2, 1.8, 2.4} { \fill[blue!60] (\i, \i) circle (1.5pt); }
        \draw[blue!80, very thick] (0.8, 1.2) -- (1.2, 1.6); 
        \draw[gray!80!black, very thick] (2.0, 0.8) -- (2.0, 1.5); 
        \node[font=\sffamily\tiny, text width=4cm] at (1.5, 0) {Quantifies consistency (DET) vs. stalling (LAM).};
    \end{scope}

    \begin{scope}[shift={(10.5,-2.5)}]
        \node[anchor=west, font=\sffamily\bfseries] at (0, 3.5) {Step 3: Temporal Signals};
        \draw[->] (0.5,0.8) -- (2.5,2.2) node[right, font=\sffamily\tiny] {Slope};
        \draw[thick, blue!60!black] (0.5, 1.5) sin (1, 2) cos (1.5, 1) sin (2, 1.5) cos (2.5, 2);
        \node[font=\sffamily\tiny, text width=4cm] at (2, 0) {Extracts DFA and trends to predict complexity/accuracy.};
    \end{scope}

    \node[anchor=west, text width=4.5cm, font=\sffamily\tiny] 
        at (-1.5, -3.5) {
        \textbf{1. Dynamical Framing:}\\
        Introduces RQA as a mechanistic tool for monitoring the dynamics of token generation.
    };
    
    \node[anchor=west, text width=4.5cm, font=\sffamily\tiny] 
        at (4.6, -3.5) {
        \textbf{2. Complexity Resolution:}\\
        Temporal signals outperform length by \textbf{8\%} in predicting task complexity.
    };
    
    \node[anchor=west, text width=4.5cm, font=\sffamily\tiny] 
        at (10, -3.5) {
        \textbf{3. Structural Indicators:}\\
        Identifies variability \textbf{DET} features and stalling acceleration \textbf{LAM} features as keys to the characterisation of the dynamics.
    };

\end{tikzpicture}
\caption{\textbf{Problems with CoT Analysis - } \textbf{Top (The Challenges):} (1) Massive token counts prevent human audit; (2) Total length is a poor proxy for reasoning quality; (3) Textual output often masks internal logic failures; (4) Static interpretability fails to see time-dependent patterns. \textbf{Bottom (RQA: the proposed solution):} We propose RQA to transform discrete tokens into a measurable \textbf{latent trajectory}, decoding the dynamics of reasoning. We demonstrate that these temporal signals significantly outperform response length in resolving task complexity and identifying structural signals of the dynamics.}
\label{fig:master_framework}
\end{figure*}
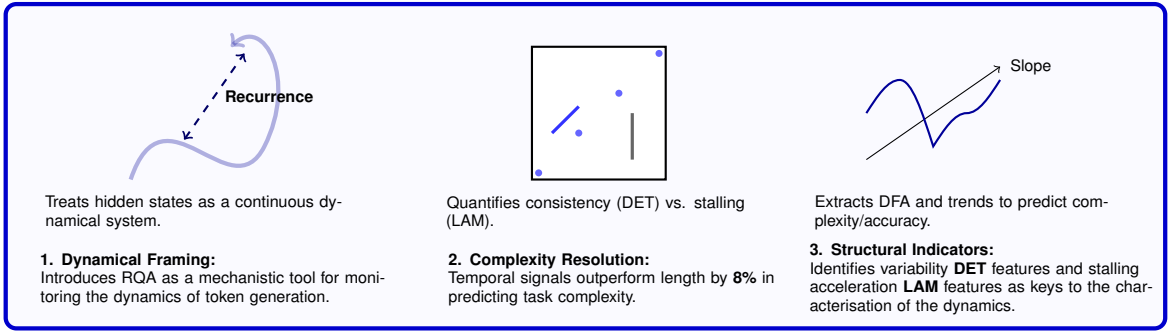

The effectiveness of test-time compute through extended "Chain-of-Thought" (CoT) sequences is central to the performance of Large Reasoning Models like Deepseek R1, or OpenAI's o3. To understand this Chain of Thought process, current research focuses on the information flow across layers \citep{yang2025exploring, ton2411understanding} or utilises continuous generative frameworks like diffusion to model CoT trajectories \citep{ye2402diffusion}. Another approach is to textually analyse and characterise linguistic behaviours of the Chain of Thought \citep{venhoff2025understanding, marjanovic2504deepseek}. On a practical level, researchers often evaluate reasoning by mapping the "Curse of Complexity" against the model’s total token output \citep{lin2025zebralogic}. These approaches aim to ground the model’s linguistic output in either its underlying computational structures or its textual behaviours. 

\textbf{Challenges:}
However, treating CoT traces as purely linguistic or static entities is increasingly impractical. \textbf{First}, the sheer volume of tokens generated during test-time scaling quickly exceeds human audit capacity, creating a verification bottleneck. \textbf{Second}, CoT traces are often prone to "overthinking," unfaithfulness, or semantic drifting, where the generated text does not strictly reflect the model's internal logic \citep{kambhampati2025stop, arcuschin2503chain, sun2025thinkedit}. \textbf{Critically}, existing mechanistic methods often treat the reasoning process as a series of static snapshots or isolated units of analysis. They fail to be effective in characterising the temporal, path-dependent dynamics of how representations evolve in the latent space, leaving a gap in our understanding of the underlying computation process of the Chain of Thought process.

\textbf{Contributions:}
In this work, we seek to reduce the reliance on direct textual analysis of Chain-of-Thought (CoT) while explicitly modelling CoT as a \emph{temporal process}. Our contributions are threefold:

\begin{itemize}[leftmargin=*, noitemsep, topsep=2pt]
    \item \textbf{Dynamical framing of CoT:}
    We model CoT generation as a dynamical process unfolding in representation space and introduce \textbf{Recurrence Quantification Analysis (RQA)} as a mechanistic interpretability framework. By treating the autoregressive generation of tokens as a dynamical system, we apply RQA to trajectories formed by the last-layer embeddings of each generated token. This maps high-dimensional latent trajectories to 2D recurrence plots, enabling the quantification of predictability (\textbf{DET}) and complexity (\textbf{ENTR}).

    \item \textbf{Empirical validation on reasoning traces:}
    We validate the proposed framework using foundational RQA measures—\textbf{Determinism (DET)}, capturing representational consistency, and \textbf{Laminarity (LAM)}, reflecting periods of semantic stalling—on 3,600 reasoning traces from \textit{DeepSeek-Distill-7B-Qwen} evaluated on the \textbf{ZebraLogic} benchmark. Temporal RQA features significantly outperform length-based baselines for task complexity classification, achieving an 8\% improvement (36.9\% vs.\ 29.0\%).

    \item \textbf{Identification of structural difficulty indicators:}
    We show that the \emph{variability of semantic repetition} and the \emph{acceleration of stalling} (Laminarity slope) are strong indicators of combinatorial difficulty.
\end{itemize}

By doing so, we establish RQA as structure-sensitive framework for monitoring reasoning models, offering a new perspective in analysing the dynamics of Chain of Thought process.

\section{Related Work}

\paragraph{Alternatives to Textual Analysis:}
Recent research has explored non-textual metrics to quantify reasoning. Information-theoretic approaches utilise entropy and bottleneck measures to map information flow across layers \citep{yang2025exploring, ton2411understanding}. Others model CoT trajectories using continuous generative frameworks, such as diffusion or flow matching, to enable self-correcting computation \citep{ye2402diffusion, yang2025policy}. While principled, these methods often require architectural modifications; RQA, by contrast, operates on the existing residual stream of any LRM.

\paragraph{Mechanistic Interpretability of Chain-of-Thought:}
There have been some Mechanistic works on analysing Chain of Thoughts \citep{tang2025enhancing, bogdan2025thought, dutta2024think}. Tang et al. \citep{tang2025enhancing} identify reasoning-critical neurons in feed-forward layers by contrasting activation patterns on high- and low-quality reasoning traces. Chen et al \citep{chen2025does}  extract latent feature directions from hidden states to study CoT faithfulness by using sparse autoencoders on the final-token activations with and without CoT. Bogdan et al. \citep{bogdan2025thought} took into account of the temporal nature of the CoT when analysing it. They used the sentence-level attribution methods for the analysis based on sentence as the grounding unit of analysis. While insightful, these methods largely either target individual neurons, features at particular layers or time steps, rather than analysing the temporal structure of a reasoning chain, or use textual sentence as the fundamental unit of analysis without addressing the potential issue of lack of semantics in LLM's CoTs \citep{kambhampati2025stop}. As a result, this leaves unaddressed the question of how internal representations evolve over time during the CoT process. This leaves a critical gap at the level of token-wise latent dynamics, where reasoning unfolds as a path-dependent process in representation space.

\paragraph{LLMs as Dynamical Systems:}
Another perspective is to view autoregressive generation through the lens of dynamical systems. Recent work has conceptualised Transformer inference as a discrete stochastic dynamical process evolving over token steps \citep{bhargava2310s, carson2025statistical}. When extended reasoning traces are generated, this process induces effective recurrence through repeated context reuse, functionally resembling recurrent computation \citep{zhang2024autoregressive+}. From this perspective, the sequence of hidden states associated with each generated token forms a high-dimensional time-series, allowing the application of RQA.

\paragraph{Recurrence Quantification Analysis:}
Recurrence is a fundamental property of dynamical systems, occurring when trajectories revisit regions of phase space \citep{eckmann1995recurrence}. RQA provides statistical measures to quantify the predictability and complexity of these dynamics without assuming linearity or stationarity \citep{webber2015recurrence}. Recently, RP-based methods have transitioned from descriptive tools in physics to components of machine learning pipelines, where time-series are analysed via geometric structure \citep{marwan2023trends}. RQA has successfully detected regime shifts in neuroscience \citep{lopes2021recurrence} and finance \citep{unal2022stability}, but has not yet been applied to the internal representations of large language models.

\section{Methodology}

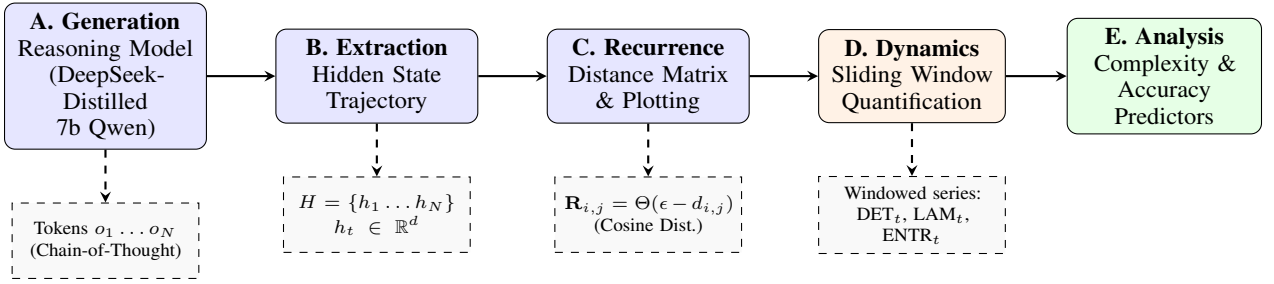
\begin{figure*}[t]
\centering
\begin{tikzpicture}[
    node distance=0.7cm and 0.9cm, 
    auto, 
    >=stealth, 
    block/.style={rectangle, draw, fill=blue!10, text width=2.4cm, text centered, rounded corners, minimum height=1.1cm, font=\small},
    data/.style={rectangle, draw, dashed, text width=2.2cm, text centered, fill=gray!5, minimum height=1.0cm, font=\scriptsize},
    process/.style={rectangle, draw, fill=orange!10, text width=2.2cm, text centered, rounded corners, minimum height=1.1cm, font=\small},
    arrow/.style={thick,->,>=stealth}
]
    \node [block] (llm) {\textbf{A. Generation} \\ Reasoning Model \\ (DeepSeek-Distilled 7b Qwen)};
    \node [data, below=of llm] (cot) {Tokens $o_1 \dots o_N$ \\ (Chain-of-Thought)};
    
    \node [block, right=of llm] (extraction) {\textbf{B. Extraction} \\ Hidden State \\ Trajectory};
    \node [data, below=of extraction] (states) {$H = \{h_1 \dots h_N\}$ \\ $h_t \in \mathbb{R}^d$};

    \node [block, right=of extraction] (geometry) {\textbf{C. Recurrence} \\ Distance Matrix \\ \& Plotting};
    \node [data, below=of geometry] (matrix) {$\mathbf{R}_{i,j} = \Theta(\epsilon - d_{i,j})$ \\ (Cosine Dist.)};

    \node [process, right=of geometry] (rqa) {\textbf{D. Dynamics} \\ Sliding Window \\ Quantification};
    \node [data, below=of rqa] (metrics) {Windowed series: \\ DET$_t$, LAM$_t$, ENTR$_t$};

    \node [rectangle, draw, fill=green!10, right=0.8cm of rqa, text width=2.3cm, text centered, font=\small, rounded corners, minimum height=1.1cm] (output) {\textbf{E. Analysis} \\ Complexity \& \\ Accuracy Predictors};

    \draw [arrow] (llm) -- (extraction);
    \draw [arrow] (extraction) -- (geometry);
    \draw [arrow] (geometry) -- (rqa);
    \draw [arrow] (rqa) -- (output);
    \draw [arrow, dashed] (llm) -- (cot);
    \draw [arrow, dashed] (extraction) -- (states);
    \draw [arrow, dashed] (geometry) -- (matrix);
    \draw [arrow, dashed] (rqa) -- (metrics);
\end{tikzpicture}
\caption{The proposed RQA interpretability pipeline. (A) Tokens are generated autoregressively. (B) Latent states form a high-dimensional trajectory. (C) Self-similarity is mapped to a recurrence matrix. (D) Non-stationary dynamics are quantified via sliding windows. (E) Temporal features (slopes, DFA) serve as inputs for downstream classification.}
\label{fig:pipeline_v2}
\end{figure*}

\subsection{LLM Token Generation as Temporal Recurrent Systems}
While Transformers are architecturally feed-forward, inference during
Chain-of-Thought (CoT) generation induces temporal dependence through
autoregressive context reuse (Fig \ref{fig:pipeline_v2}, \textbf{Block A}). At each step $t$, the model generates a
token, which is appended to the input sequence and re-embedded at
step $t{+}1$. For each generated token, we extract the corresponding
final-layer hidden state $h_t \in \mathbb{R}^d$ (\textbf{Block B}). The ordered sequence
\[
\mathcal{T} = \{h_1, h_2, \ldots, h_N\}
\]
defines a latent trajectory whose evolution is shaped by repeated
reuse of prior outputs. Although no explicit recurrent state is
maintained \citep{zhang2024autoregressive+}, this autoregressive loop \emph{effectively approximates
recurrent dynamics}, inducing structured temporal dependencies in the
latent representation space.

\begin{figure*}[t]
\vspace{-0.4em}
\centering
\includegraphics[width=0.46\textwidth]{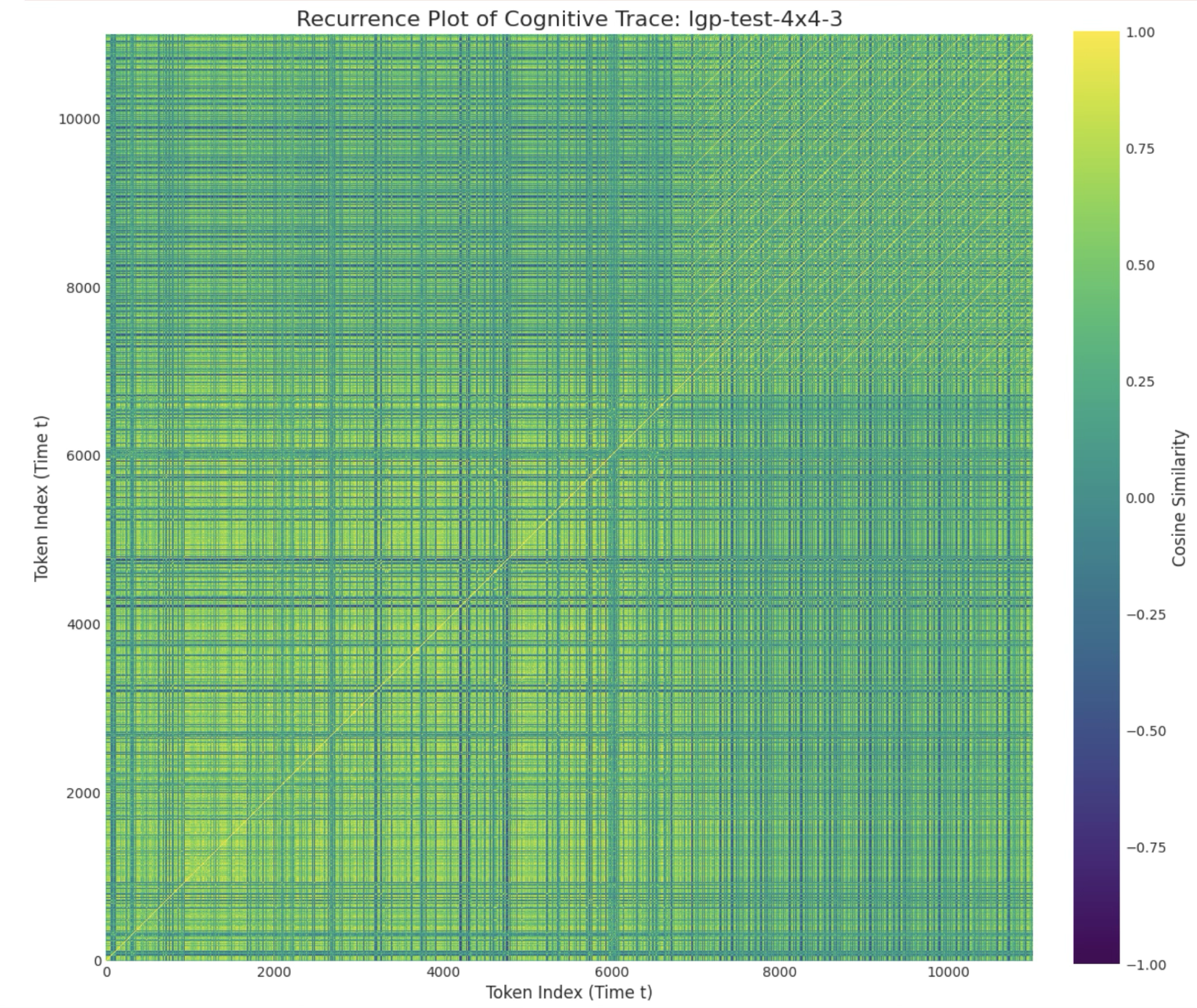}\hfill
\includegraphics[width=0.43\textwidth]{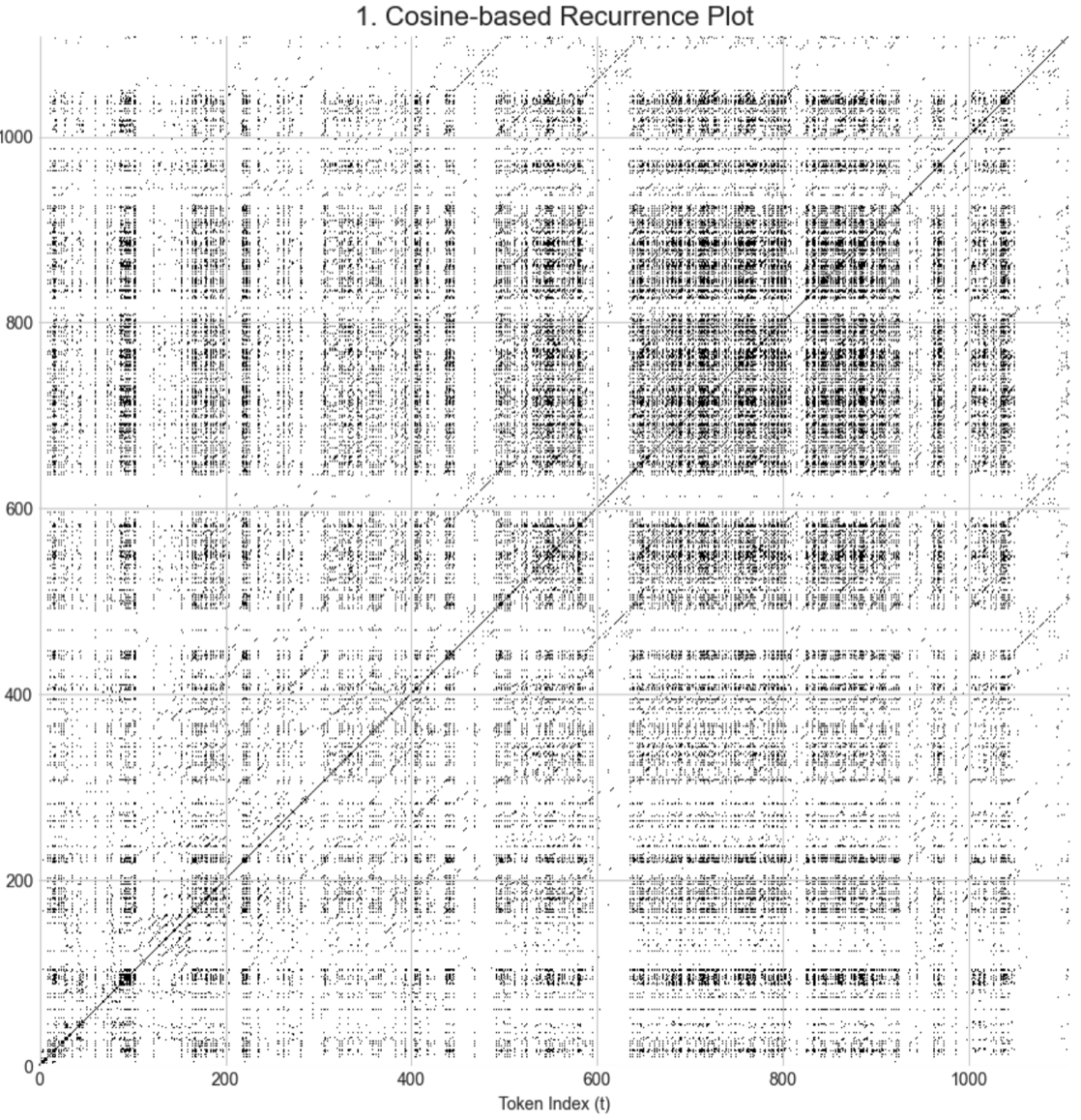}
\vspace{-0.3em}
\caption{\small
Hidden-state trajectory visualisation.
Cosine similarity and recurrence plot ($\epsilon=0.1$) illustrating DET and LAM
structures.
}
\vspace{-0.6em}
\label{fig:trajectory_analysis}
\end{figure*}

\subsection{Recurrence Rate}
By treating intermediate token generation as a time-ordered, high-dimensional
trajectory $\mathcal{T} = \{ \mathbf{h}_1, \mathbf{h}_2, \ldots, \mathbf{h}_N \}$,
we analyse reasoning dynamics through \emph{Recurrence Quantification Analysis}
(RQA). The core object of RQA is the \textbf{recurrence matrix}, which encodes
pairwise revisitations of the latent state space.

Given a distance norm $d(\cdot,\cdot)$ on $\mathbb{R}^d$ and a recurrence
threshold $\varepsilon$, the recurrence matrix $\mathbf{R} \in \{0,1\}^{N \times N}$
is defined element-wise as
\begin{equation}
R_{i,j} = \Theta\!\left( \varepsilon - d(\mathbf{h}_i, \mathbf{h}_j) \right),
\end{equation}
where $\Theta(\cdot)$ denotes the Heaviside step function. In our experiments,
$d(\cdot,\cdot)$ is instantiated as cosine distance.

This construction can be viewed as the application of a non-linear operator
$\mathcal{G}_\varepsilon$ that maps a high-dimensional trajectory to a binary
recurrence structure (\textbf{Block C}):
\[
\mathbf{R} = \mathcal{G}_\varepsilon(\mathcal{T}).
\]
The resulting \textbf{Recurrence Plot} is the graphical visualisation of
$\mathbf{R}$, where a point is plotted at coordinates $(i,j)$ if and only if
$R_{i,j}=1$. This transformation projects the latent trajectory into a
two-dimensional representation that preserves the semantic
structure of the underlying dynamics.

Figure~\ref{fig:trajectory_analysis} illustrates the result of the process, showing the
cosine-similarity matrix, its binarisation using a $10^{\text{th}}$ percentile
threshold, and the resulting recurrence structures.

\subsection{Quantification Metrics}Figure~\ref{fig:pipeline_v2} demonstrates the extraction process, aimining to characterise the dynamics of the representation token generation process. Following the binarisation, we apply RQA to extract structural signatures (\textbf{Block D}). Unlike textual analysis, RQA measures the \textit{geometry} of the computation: Here, in accordance with common practices \citep{webber2015recurrence}, we exclude the main diagonal ($i=j$) from all calculations to avoid inflationary signal.

\paragraph{Determinism (DET):}
DET measures the proportion of recurrence points that form diagonal lines of at least length $l_{\min}=3$:
\begin{equation}
\label{eq:det}
\mathrm{DET} = 
\frac{\sum_{l=l_{\min}}^{N} l\,P(l)}
     {\sum_{i,j \neq i}^{N} R_{i,j}},
\end{equation}
where $P(l)$ denotes the histogram of diagonal line lengths.
Geometrically, diagonal lines correspond to parallel trajectories in phase space, indicating \textbf{predictability and structural consistency} \citep{marwan2023trends}. 
In our context, high DET serves as an index of \textbf{semantic repetition}, reflecting the execution of stable and predictable representations.

\paragraph{Laminarity (LAM):}
LAM quantifies the proportion of recurrence points forming vertical or horizontal lines of minimum length $v_{\min}=3$:
\begin{equation}
\label{eq:lam}
\mathrm{LAM} =
\frac{\sum_{v=v_{\min}}^{N} v\,P(v)}
     {\sum_{i,j \neq i}^{N} R_{i,j}},
\end{equation}
where $P(v)$ is the histogram of vertical line lengths.
LAM captures \textbf{laminar regimes} in which the system remains confined to a localised region of state space \citep{webber2015recurrence}. 
We interpret high LAM as \textbf{semantic stalling}, indicating that the model stays in a static semantic configuration (e.g., repeatedly re-evaluating constraints) without substantive representation progression.

\paragraph{Recurrence Entropy (ENTR):}
ENTR is defined as the Shannon entropy of the probability distribution of diagonal line lengths $p(l)$:
\begin{equation}
\label{eq:entr}
\mathrm{ENTR} = -\sum_{l=l_{\min}}^{N} p(l)\,\ln p(l).
\end{equation}
ENTR measures the \textbf{diversity of the deterministic structure} \citep{webber2015recurrence}. 
High ENTR reflects a heterogeneous repertoire of reasoning routines, characteristic of deep, multi-stage search processes, whereas low ENTR indicates either repetitive loops (stalling) or a single dominant reasoning trajectory.

\paragraph{Temporal Dynamics:}
As LLM reasoning dynamics are inherently \textbf{non-stationary} \citep{marwan2023trends}, we compute RQA metrics over successive sliding windows ($W=150$, step size $=15$). 
From each resulting metric time series (e.g., $\mathrm{DET}_t$), we extract the \textbf{detrended fluctuation analysis (DFA)} \citep{peng1994mosaic} scaling exponent and the linear trend slope. 
These temporal descriptors capture transitions of model latent representations throughout the token generation process.

\section{Experiment}
\label{sec:experiment}

\subsection{Dataset and Task: ZebraLogic}

When evaluating model's performance, we seek to examine its logical capability without being bias by its knowledge, and a structured dataset that allows use to examine model's behaviours across increasing levels of difficulty. To this end, we use the \textbf{ZebraLogic} benchmark \citep{lin2025zebralogic}, an evaluation suite designed to assess the logical reasoning capabilities of Large Language Models (LLMs) through synthetic logic-grid puzzles. Each instance is formulated as a \textbf{Constraint Satisfaction Problem (CSP)} consisting of $N$ houses and $M$ features per house. Each feature takes exactly $N$ possible values, and the model must infer a unique assignment for every feature--house pair based on a set of linguistic clues.

Task difficulty is grounded in the factorial growth of the underlying combinatorial search space. For an $N \times M$ puzzle, the number of possible assignments prior to applying any constraints is:
\begin{equation}
\label{eq:search_space}
\text{Search Space Size} = (N!)^M
\end{equation}
This factorial dependence induces a rapid combinatorial explosion as grid dimensions increase, providing a rigorous, language-independent measure of task complexity.

\subsection{Data Collection and Preliminary Performance}
\begin{table}[h]
\centering
\small
 \begin{tabular}{lcccc}
\toprule
\textbf{Difficulty} & \textbf{Incorrect} & \textbf{Correct} & \textbf{Total} & \textbf{Mean Acc.} \\
\midrule
2$\times$3 & 25 & 375 & 400 & 0.9375 \\
2$\times$4 & 85 & 315 & 400 & 0.7875 \\
2$\times$5 & 147 & 253 & 400 & 0.6325 \\
3$\times$2 & 57 & 343 & 400 & 0.8575 \\
3$\times$3 & 179 & 221 & 400 & 0.5525 \\
3$\times$4 & 276 & 124 & 400 & 0.3100 \\
4$\times$3 & 310 & 90 & 400 & 0.2250 \\
4$\times$4 & 391 & 9 & 400 & 0.0225 \\
5$\times$2 & 316 & 84 & 400 & 0.2100 \\
\midrule
\textbf{Totals} & \textbf{1786} & \textbf{1814} & \textbf{3600} & \textbf{0.5039} \\
\bottomrule
\end{tabular}
\caption{Final accuracy distribution across the 3,600 samples generated for RQA analysis, categorised by puzzle configuration.}
\label{tab:merged_accuracy}
\end{table}
\begin{figure}[h]
\centering
\includegraphics[width=0.9\linewidth]{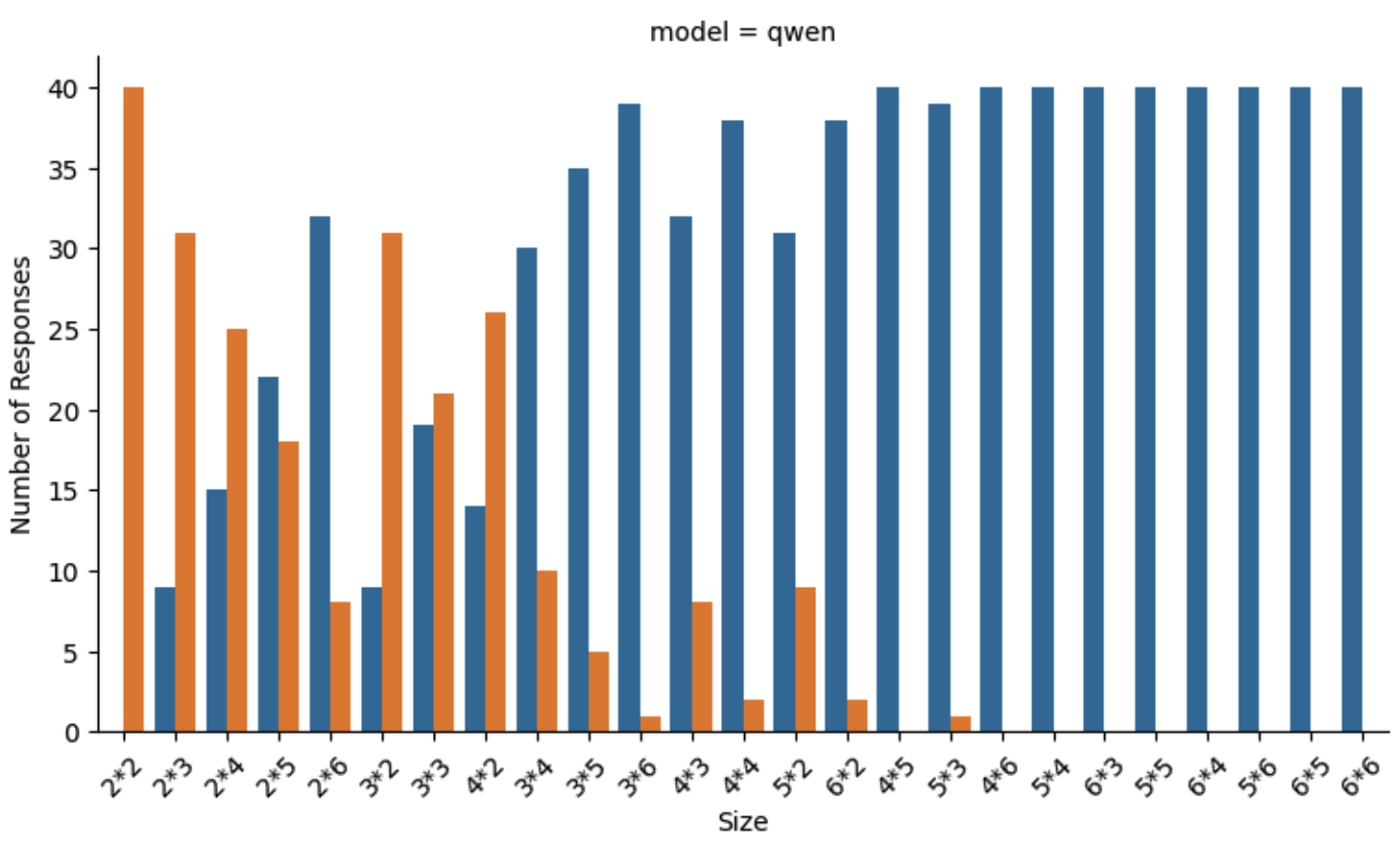}
\caption{Preliminary evaluation showing the inverse correlation between combinatorial complexity and puzzle-level accuracy across DeepSeek distilled models.}
\label{fig:accuracy_trend}
\end{figure}

We generated a corpus of reasoning trajectories using \textbf{DeepSeek-R1-Distill-Qwen-7B}, selected for its ability to produce explicit Chain-of-Thought (CoT) reasoning tokens. Following the developer’s recommendations for eliciting robust reasoning behavior, we employed stochastic decoding with a temperature of $0.6$.

\paragraph{Preliminary Analysis:}
Prior to conducting Recurrence Quantification Analysis (RQA), we observed an \textbf{negative correlation with complexity}: in line with the ``Curse of Complexity'' \citep{lin2025zebralogic}, model accuracy decreases sharply as the combinatorial search space grows (Figure~\ref{fig:accuracy_trend}).

\paragraph{Data Selection and Class Balancing:}
To obtain a balanced distribution of correct and incorrect reasoning traces, we selected nine grid configurations: 

$2\times 3,~2\times 4,~2\times 5,~3\times 2,~3\times 3,~3\times 4,~4\times 3,~4\times 4,~5\times 2.$

These configurations were chosen for the following reasons:

\begin{itemize}[leftmargin=*, noitemsep]
    \item \textbf{Class balancing for correctness:} In larger configurations (e.g., $6\times6$), the model’s accuracy approaches zero. Including such extreme cases would create a ``floor effect,'' making binary classification of correctness statistically unreliable. The nine selected sizes provide a near-perfect global balance of 1,814 correct versus 1,786 incorrect traces (Table~\ref{tab:merged_accuracy}).
    \item \textbf{Representativeness:} The selected configurations span the complexity spectrum—from $2\times3$ (low search space, high accuracy) to $4\times4$ and $5\times2$ (high search space, low accuracy). This allows us to capture the effects of the ``Curse of Complexity'' without introducing data sparsity issues associated with the most extreme grid sizes.
\end{itemize}

For each configuration, we sampled 40 distinct puzzles and generated 10 independent reasoning traces per puzzle, yielding a total of 3,600 traces.

\subsection{Feature Extraction: Recurrence Quantification Analysis}
To characterise the internal dynamics of the model’s reasoning process, we extracted hidden-state trajectories from the final transformer layer for each reasoning trace. From these trajectories, we derived three class of features for comparisation:

\paragraph{1. Response Length:}
The total number of tokens in the generated CoT response.

\paragraph{2. Global RQA:}
Full-trace averages of Determinism (DET), Laminarity (LAM), and Recurrence Entropy (ENTR), capturing the overall recurrence structure of the hidden-state dynamics.

\paragraph{3. Temporal RQA:}
To model non-stationary reasoning dynamics, we applied a sliding window of 150 tokens with a 10\% step size across each trace. For each RQA metric, we computed the mean, standard deviation, linear trend (slope), and the Detrended Fluctuation Analysis (DFA) exponent of the resulting time series.

\paragraph{RQA Hyperparameters:}
To ensure consistency across varying trace lengths and to avoid spurious recurrences, we fixed the following parameters throughout all experiments:
\begin{itemize}[leftmargin=*, noitemsep]
    \item \textbf{Recurrence threshold ($\epsilon$):} Set to the top 10th percentile of pairwise cosine distances between hidden states.
    \item \textbf{Minimum line lengths ($l_{\min}, v_{\min}$):} Fixed at 3 for diagonal and vertical lines.
    \item \textbf{Sequence cap:} Reasoning traces were truncated at 32,000 tokens.
\end{itemize}

\subsection{Classification Experiment Design}
We evaluate whether recurrence-based features provide predictive signals for two classification tasks:
\begin{itemize}[leftmargin=*, noitemsep]
    \item \textbf{Problem complexity:} A 9-way classification task predicting the grid configuration ($N \times M$).
    \item \textbf{Answer correctness:} A binary classification task distinguishing correct from incorrect solutions.
\end{itemize}

We employ two classifier families: \textbf{Logistic Regression (LR)} to assess linear separability, and \textbf{Random Forests (RF)} to capture non-linear interactions among RQA features. Specifically, we used 100 n-estimators for the Random Forests Classifier. 

\paragraph{Evaluation Protocol:}
To prevent leakage of puzzle-specific information, we use stratified 8-fold group cross-validation, where the \textbf{Puzzle ID} serves as the grouping variable. This ensures that reasoning traces derived from the same puzzle never appear in both training and test splits, enforcing strict generalisation.

\section{Results and Discussion}
\label{sec:resultsanddiscussion}

Our evaluation reveals a clear functional separation between the predictive power of response length and RQA-based methods.

\begin{figure*}[t]
  \includegraphics[width=0.48\linewidth]{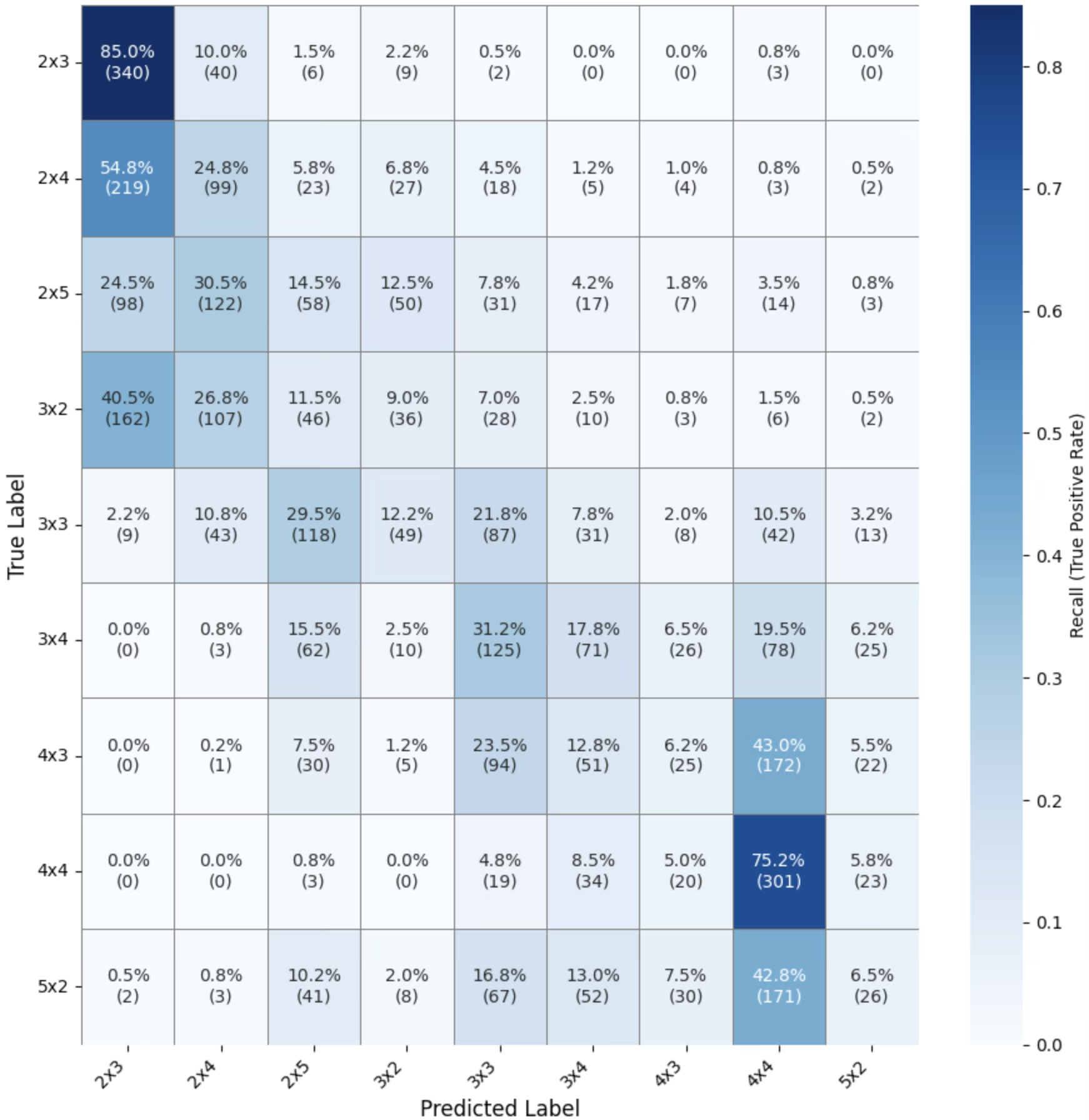} \hfill
  \includegraphics[width=0.48\linewidth]{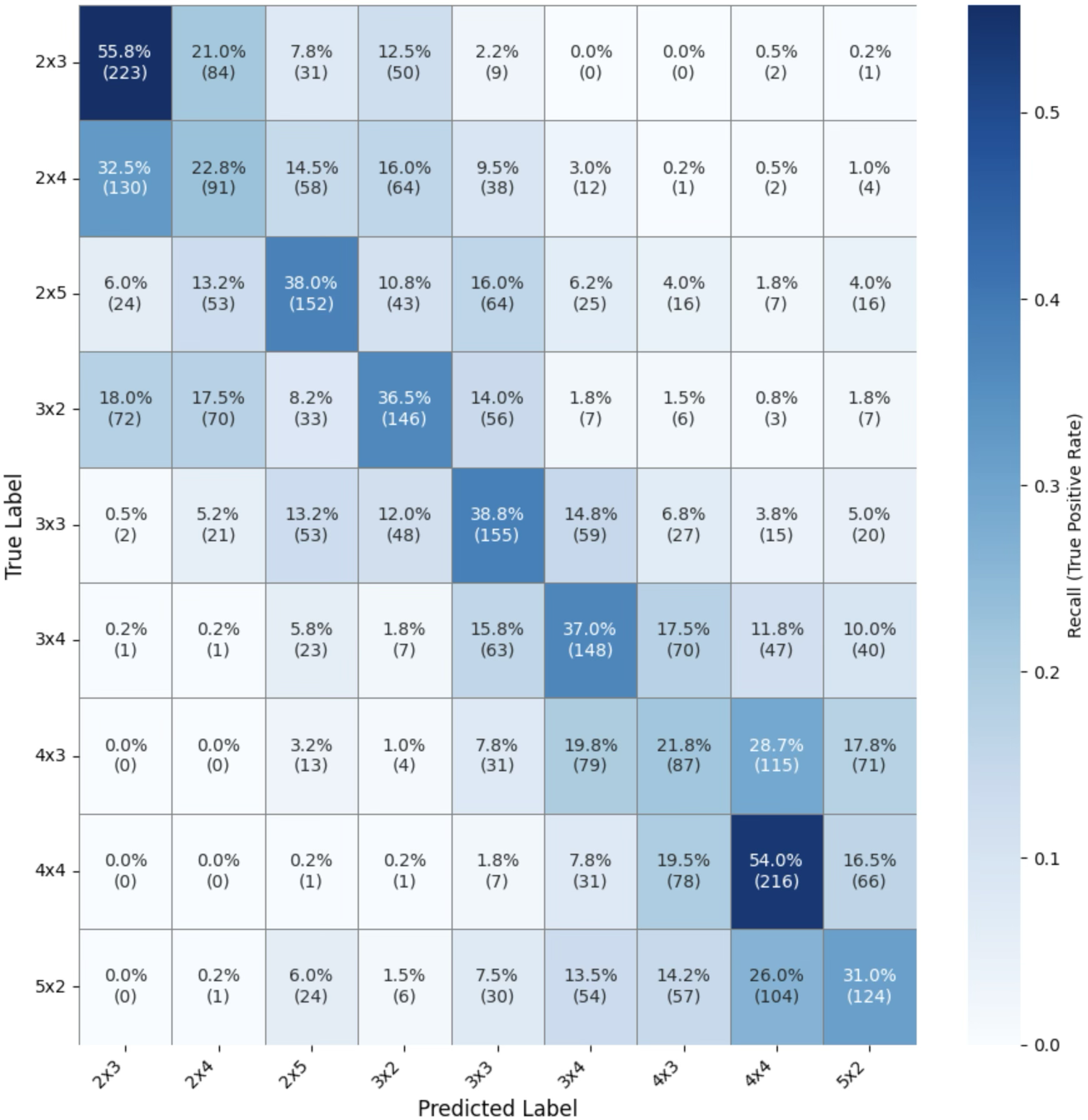}
  \caption{
    Confusion matrices for complexity classification.
    (\textbf{a}) Response-length baseline, which performs well on extreme complexity levels but struggles in intermediate regimes.
    (\textbf{b}) Temporal RQA, which exhibits more uniform performance across complexity levels.
    }
  \label{fig:cm_comparison}
\end{figure*}

\begin{table*}[t]
\centering
\small
\begin{tabular}{@{}llcc@{}}
\toprule
\textbf{Features} & \textbf{Model} & \textbf{Complexity (\%)} & \textbf{Accuracy (\%)} \\
\midrule
Baseline (Length) & RF / LR & $28.97 \pm 3.1$ & $85.58 \pm 1.5$ \\
Global RQA        & RF      & $16.81 \pm 2.2$ & $55.03 \pm 3.4$ \\
Temporal RQA      & LR      & $29.50 \pm 3.8$ & $81.19 \pm 2.0$ \\
\textbf{Temp.\ RQA} & \textbf{RF} & $\mathbf{36.94 \pm 3.3}$ & $85.00 \pm 1.2$ \\
\bottomrule
\end{tabular}
\caption{Classification results using 8-fold group cross-validation.
Temporal RQA captures task complexity where length fails, while length remains a
dominant proxy for binary failure.}
\label{tab:results_main}
\end{table*}

\paragraph{Complexity: Structural Dynamics Beyond Token Length.}
As shown in Table~\ref{tab:results_main}, the \textbf{Temporal RQA (RF)} model
substantially outperforms the length-based baseline by 8\% for complexity classification
(36.94\% vs.\ 28.97\%).
McNemar’s test confirms that this 8\% absolute improvement is statistically
significant ($p<0.05$) in 7 out of 8 validation folds (average $p<0.005$).
While response length reflects the \emph{duration} of test-time scaling, it remains
blind to the \emph{structural quality} of the underlying search process.
In contrast, the superior performance of Temporal RQA demonstrates that
combinatorial difficulty is encoded in the geometric \emph{shape} of the reasoning
trajectory.
The effectiveness of the RF classifier further suggests that task complexity emerges
as a non-linear interaction between structural stability (DET) and semantic
stalling (LAM), rather than as a monotonic function of token count.

\paragraph{Accuracy: Overlapping Signals for Binary Failure.}
For binary accuracy prediction, response length remains a remarkably strong
predictor (85.58\%), consistent with prior observations that failure often manifests
as uncontrolled trace expansion—a regime termed the \textbf{Curse of Complexity}
\citep{lin2025zebralogic}.
Temporal RQA achieves comparable performance (85.00\%), with no statistically
significant difference in the majority of folds ($p>0.05$).
This indicates that while length and RQA capture distinct aspects of the dynamics, they encode partially overlapping signals for coarse-grained failure
detection.
In particular, extreme breakdowns in reasoning dynamics are often accompanied
by both prolonged generation and increased recurrence.

\paragraph{Resolution and Non-Stationarity.}
The poor performance of Global RQA (16.81\%) highlights the fundamentally
\textbf{non-stationary} nature of LLM reasoning.
Averaging dynamics over the entire trace obscures critical transitions between
problem parsing, active logical inference, and final decoding.
Predictive signal resides almost entirely in the \emph{temporal evolution} of the
trajectory.
Consistent with this interpretation, Figure~\ref{fig:cm_comparison} shows that the length baseline performs well only at
the extremes of task difficulty, whereas Temporal RQA maintains discriminative
power across the full complexity spectrum—resolving intermediate combinatorial
regimes that length alone fails to separate.

\section{Ablation Study} 
\begin{figure}[H]
    \centering
    \includegraphics[width=1\linewidth]{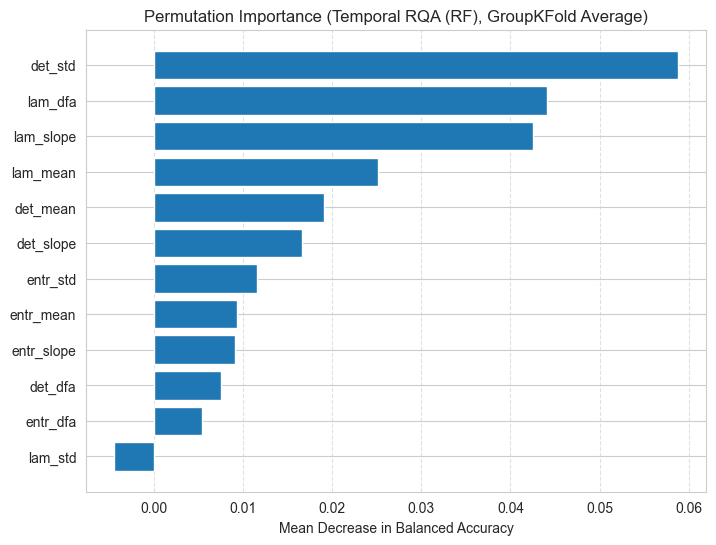}
    \caption{Here, det stands for determinism, entr is entropy, and lam is laminarity. The plot shows different contributions of each factor to the predictive power of Temporal RQA model.}
    \label{fig:perm_importance}
\end{figure}
To identify the drivers of complexity prediction, we used the Permutation Importance (Mean Decrease in Balanced Accuracy) averaged across 8-fold GroupKFold validation to help us analyse the model. As shown in Figure~\ref{fig:perm_importance}, the results reveal a distinct hierarchy of dynamical signals: 

\begin{itemize}[leftmargin=*, noitemsep, topsep=2pt]
    \item \textbf{Fluctuation of Consistency:} Most important feature is \textbf{det\_std} (variability of semantic repetition). Complexity is captured not by the stability of the amount of repetition, but by its \textit{fluctuation}. Complex puzzles induce a dynamics alternating between structured semantic repetition and its lack thereof.
    
    \item \textbf{Long-range Stalling Dynamics:} High-ranking features \textbf{lam\_dfa} (fractal memory of stalling) and \textbf{lam\_slope} (acceleration of stalling) validate Temporal RQA. The DFA exponent shows reasoning difficulty involves long-range correlations, where the model’s current “stalling” behaviour depends on representations of hundreds of steps prior.
    
    \item \textbf{Mean States:} \textbf{lam\_mean} is less important than fluctuation and memory features (\textbf{det\_std}, \textbf{lam\_dfa}). While average stalling correlates with puzzle length, RQA’s discriminative power comes from higher-order temporal rhythms rather than simple averages.
\end{itemize}

\section{Conclusion}
We introduced Recurrence Quantification Analysis (RQA) as a complementary
framework for analysing chain of thought embedding layers' object.
By analysing hidden-state trajectories as dynamical systems, RQA provides a
non-linguistic, structure-sensitive view of reasoning that operates alongside
textual explanations and scalar proxies such as response length.
Our results show that task complexity is encoded in the temporal evolution of
latent dynamics rather than in token volume alone, while binary failure remains
well captured by length variable. Our ablation study reveals an interesting dynamics between Laminarity and determinism variables in the complexity metrics that warrants further investigations into fully elaborating their relationships. As a non-textual tool, RQA potentially enables automated
monitoring during inference.
In particular, reasoning models could implement \emph{early-exit} or
\emph{backtracking} triggers, ensuring better performance and efficiency.
Together, these findings position RQA as a principled tool for probing the
geometry of reasoning processes and highlight the value of dynamical analyses
for understanding, monitoring, and controlling large-scale inference.

\section*{Limitations}

While Recurrence Quantification Analysis (RQA) offers a high-resolution,
structure-sensitive view of latent reasoning dynamics, it incurs higher
computational cost than surface-level proxies such as response length.
Computing recurrence matrices and temporal RQA features scales
quadratically with sequence length and requires access to hidden-state
activations, which may limit applicability in latency- or
resource-constrained settings.

Our methodology currently relies on heuristic choices for key RQA
hyperparameters, including the recurrence threshold (set to 10\%) and
sliding window size.
Although these parameters are held constant across experiments,
dynamical systems analyses can be sensitive to such choices.
Future work should explore different but robust, automated parameter selection
strategies to improve
stability and consistency across tasks and models.

The scope of our empirical evaluation is also limited.
We focus on a single family of reasoning models (DeepSeek-R1-Distill) and
a controlled symbolic reasoning benchmark (ZebraLogic), which provides a
clear measure of combinatorial complexity but represents a narrow task
distribution.
Establishing the generality of recurrence-based signatures will require
extending this analysis to a broader range of architectures (e.g.,
LLaMA~3, Claude) and domains, including mathematical reasoning, code
generation, and open-ended language tasks.

Finally, our analysis is restricted to hidden-state trajectories from the
final transformer layer.
While this choice offers a consistent and interpretable starting point,
recurrence structure may vary across layers.
Extending RQA to multi-layer or cross-layer dynamics, as well as
systematically analysing sensitivity to architectural depth, remains an
important direction for future research.

\section*{Acknowledgments}
We would like to acknowledge the support of the ARC Centre of Excellence for Automated Decision Making and Society (CE200100005).

\bibliography{custom}

@article{peng1994mosaic,
  title={Mosaic organization of DNA nucleotides},
  author={Peng, C-K and Buldyrev, Sergey V and Havlin, Shlomo and Simons, Michael and Stanley, H Eugene and Goldberger, Ary L},
  journal={Physical review e},
  volume={49},
  number={2},
  pages={1685},
  year={1994},
  publisher={APS}
}

@article{venhoff2025understanding,
  title={Understanding reasoning in thinking language models via steering vectors},
  author={Venhoff, Constantin and Arcuschin, Iv{\'a}n and Torr, Philip and Conmy, Arthur and Nanda, Neel},
  journal={arXiv preprint arXiv:2506.18167},
  year={2025}
}

@article{marjanovic2504deepseek,
  title={Deepseek-r1 thoughtology: Let’s think about llm reasoning},
  author={Marjanovic, Sara Vera and Patel, Arkil and Adlakha, Vaibhav and Aghajohari, Milad and BehnamGhader, Parishad and Bhatia, Mehar and Khandelwal, Aditi and Kraft, Austin and Krojer, Benno and Lu, Xing Han and others},
  journal={URL https://arxiv. org/abs/2504.07128},
  year={2025}
}

@article{dutta2024think,
  title={How to think step-by-step: A mechanistic understanding of chain-of-thought reasoning},
  author={Dutta, Subhabrata and Singh, Joykirat and Chakrabarti, Soumen and Chakraborty, Tanmoy},
  journal={arXiv preprint arXiv:2402.18312},
  year={2024}
}

@article{bogdan2025thought,
  title={Thought Anchors: Which LLM Reasoning Steps Matter?},
  author={Bogdan, Paul C and Macar, Uzay and Nanda, Neel and Conmy, Arthur},
  journal={arXiv preprint arXiv:2506.19143},
  year={2025}
}

@article{chen2025does,
  title={How does chain of thought think? mechanistic interpretability of chain-of-thought reasoning with sparse autoencoding},
  author={Chen, Xi and Plaat, Aske and van Stein, Niki},
  journal={arXiv preprint arXiv:2507.22928},
  year={2025}
}

@inproceedings{tang2025enhancing,
  title={Enhancing Chain-of-Thought Reasoning via Neuron Activation Differential Analysis},
  author={Tang, Yiru and Zhou, Kun and Min, Yingqian and Zhao, Wayne Xin and Sha, Jing and Sheng, Zhichao and Wang, Shijin},
  booktitle={Proceedings of the 2025 Conference on Empirical Methods in Natural Language Processing},
  pages={16162--16170},
  year={2025}
}

@article{yang2025policy,
  title={Policy-to-Language: Train LLMs to Explain Decisions with Flow-Matching Generated Rewards},
  author={Yang, Xinyi and Zeng, Liang and Dong, Heng and Yu, Chao and Wu, Xiaoran and Yang, Huazhong and Wang, Yu and Tambe, Milind and Wang, Tonghan},
  journal={arXiv preprint arXiv:2502.12530},
  year={2025}
}

@article{ye2402diffusion,
  title={Diffusion of Thoughts: Chain-of-Thought Reasoning in Diffusion Language Models, December 2024},
  author={Ye, Jiacheng and Gong, Shansan and Chen, Liheng and Zheng, Lin and Gao, Jiahui and Shi, Han and Wu, Chuan and Jiang, Xin and Li, Zhenguo and Bi, Wei and others},
  journal={arXiv preprint arXiv:2402.07754},
  year={2024}
}

@article{yang2025exploring,
  title={Exploring Information Processing in Large Language Models: Insights from Information Bottleneck Theory},
  author={Yang, Zhou and Qi, Zhengyu and Ren, Zhaochun and Jia, Zhikai and Sun, Haizhou and Zhu, Xiaofei and Liao, Xiangwen},
  journal={arXiv preprint arXiv:2501.00999},
  year={2025}
}

@article{ton2411understanding,
  title={Understanding chain-of-thought in llms through information theory, 2024},
  author={Ton, Jean-Francois and Taufiq, Muhammad Faaiz and Liu, Yang},
  journal={URL https://arxiv. org/abs/2411.11984},
  year={2024}
}

@article{kambhampati2025stop,
  title={Stop Anthropomorphizing Intermediate Tokens as Reasoning/Thinking Traces!},
  author={Kambhampati, Subbarao and Stechly, Kaya and Valmeekam, Karthik and Saldyt, Lucas and Bhambri, Siddhant and Palod, Vardhan and Gundawar, Atharva and Samineni, Soumya Rani and Kalwar, Durgesh and Biswas, Upasana},
  journal={arXiv preprint arXiv:2504.09762},
  year={2025}
}

@article{marwan2023trends,
  title={Trends in recurrence analysis of dynamical systems},
  author={Marwan, Norbert and Kraemer, K Hauke},
  journal={The European Physical Journal Special Topics},
  volume={232},
  number={1},
  pages={5--27},
  year={2023},
  publisher={Springer}
}

@article{sun2025thinkedit,
  title={ThinkEdit: Interpretable Weight Editing to Mitigate Overly Short Thinking in Reasoning Models},
  author={Sun, Chung-En and Yan, Ge and Weng, Tsui-Wei},
  journal={arXiv preprint arXiv:2503.22048},
  year={2025}
}

@article{arcuschin2503chain,
  title={Chain-of-thought reasoning in the wild is not always faithful, 2025},
  author={Arcuschin, Iv{\'a}n and Janiak, Jett and Krzyzanowski, Robert and Rajamanoharan, Senthooran and Nanda, Neel and Conmy, Arthur},
  journal={URL https://arxiv. org/abs/2503.08679},
  year={2025}
}

@article{zhang2024autoregressive+,
  title={Autoregressive+ Chain of Thought= Recurrent: Recurrence's Role in Language Models' Computability and a Revisit of Recurrent Transformer},
  author={Zhang, Xiang and Abdul-Mageed, Muhammad and Lakshmanan, Laks VS},
  journal={arXiv preprint arXiv:2409.09239},
  year={2024}
}

@incollection{eckmann1995recurrence,
  title={Recurrence plots of dynamical systems},
  author={Eckmann, J-P and Kamphorst, S Oliffson and Ruelle, David},
  booktitle={Turbulence, strange attractors and chaos},
  pages={441--445},
  year={1995},
  publisher={World Scientific}
}

@article{webber2015recurrence,
  title={Recurrence quantification analysis},
  author={Webber, Charles L and Marwan, Norbert},
  journal={Theory and best practices},
  volume={426},
  year={2015},
  publisher={Springer}
}

@article{lopes2021recurrence,
  title={Recurrence quantification analysis of dynamic brain networks},
  author={Lopes, Marinho A and Zhang, Jiaxiang and Krzemi{\'n}ski, Dominik and Hamandi, Khalid and Chen, Qi and Livi, Lorenzo and Masuda, Naoki},
  journal={European Journal of Neuroscience},
  volume={53},
  number={4},
  pages={1040--1059},
  year={2021},
  publisher={Wiley Online Library}
}

@article{carson2025statistical,
  title={A Statistical Physics of Language Model Reasoning},
  author={Carson, Jack David and Reisizadeh, Amir},
  journal={arXiv preprint arXiv:2506.04374},
  year={2025}
}

@article{bhargava2310s,
  title={What’s the magic word? A control theory of LLM prompting (2023)},
  author={Bhargava, Aman and Witkowski, Cameron and Shah, M and Thomson, MW},
  journal={URL https://arxiv.org/abs/2310.04444},
  year={2023}
}

@article{lin2025zebralogic,
  title={Zebralogic: On the scaling limits of llms for logical reasoning},
  author={Lin, Bill Yuchen and Bras, Ronan Le and Richardson, Kyle and Sabharwal, Ashish and Poovendran, Radha and Clark, Peter and Choi, Yejin},
  journal={arXiv preprint arXiv:2502.01100},
  year={2025}
}

@article{unal2022stability,
  title={Stability analysis of bitcoin using recurrence quantification analysis},
  author={{\"U}nal, Baki},
  journal={Chaos Theory and Applications},
  volume={4},
  number={2},
  pages={104--110},
  year={2022},
  publisher={Akif AKG{\"U}L}
}

\appendix

\end{document}